# Localizing and Orienting Street Views Using Overhead Imagery

Nam N. Vo and James Hays

Georgia Institute of Technology
{namvo,hays}@gatech.edu

**Abstract.** In this paper we aim to determine the location and orientation of a ground-level query image by matching to a reference database of overhead (e.g. satellite) images. For this task we collect a new dataset with one million pairs of street view and overhead images sampled from eleven U.S. cities. We explore several deep CNN architectures for cross-domain matching – Classification, Hybrid, Siamese, and Triplet networks. Classification and Hybrid architectures are accurate but slow since they allow only partial feature precomputation. We propose a new loss function which significantly improves the accuracy of Siamese and Triplet embedding networks while maintaining their applicability to large-scale retrieval tasks like image geolocalization. This image matching task is challenging not just because of the dramatic viewpoint difference between ground-level and overhead imagery but because the orientation (i.e. azimuth) of the street views is unknown making correspondence even more difficult. We examine several mechanisms to match in spite of this – training for rotation invariance, sampling possible rotations at query time, and explicitly predicting relative rotation of ground and overhead images with our deep networks. It turns out that explicit orientation supervision *also* improves location prediction accuracy. Our best performing architectures are roughly 2.5 times as accurate as the commonly used Siamese network baseline.

## 1 Introduction

In this work we propose deep learning approaches to the problem of ground to overhead image matching. Such approaches enable large scale image geolocalization techniques to use widely-available overhead/satellite imagery to estimate the location of ground level photos. This is in contrast to typical image geolocalization which relies on matching "ground-to-ground" using a reference database of geotagged photographs. It is comparatively easy (for humans and machines) to determine if two ground level photographs depict the same location, but the world is very non-uniformly sampled by tourists and street-view vehicles. On the other hand, overhead imagery densely covers the Earth thanks to satellites and other aerial surveys. Because of this widespread coverage, matching ground-level photos to overhead imagery has become an attractive geolocalization approach [19]. However, it is a very challenging task (even for humans) because of the huge viewpoint variation and often lighting and seasonal variations, too. In this paper we try to learn how to match urban and suburban images from street-view to overhead-view imagery at fine-scale. As shown in Figure 1, once the matching is done, the results can be ranked to generate a location estimate for a ground-level query.

To address cross-view geolocalization, the community has recently found deep learning techniques to outperform hand-crafted features [20, 34]. These approaches adopt architectures from the similar task of face verification [7, 29]. The method is as follows: a CNN, more specifically a Siamese architecture network [5, 7], is used to learn a common low dimensional feature representation for both ground level and aerial image, where they can be compared to determine a matching score. While being superior to non-deep approaches (or pre-trained deep features), we show there is significant room for improvement.

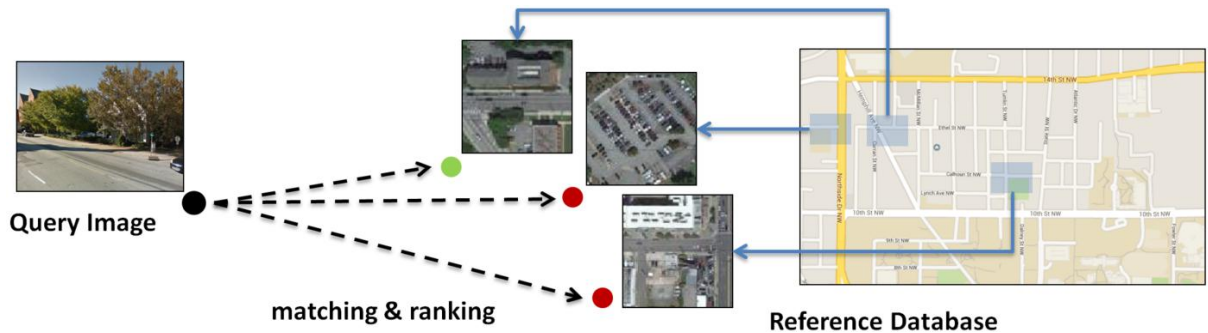

**Fig. 1.** Street-view to overhead-view image matching

To that end we study different deep learning approaches for matching/verification and ranking/retrieval tasks. We develop better loss functions using the novel distance based logistic (DBL) layer. To further improve the performance, we show that good representations can be learned by incorporating rotational invariance (RI) and orientation regression (OR) during training. Experiments are performed on a new large scale dataset which will be published to encourage future research. We believe the findings here generalize to similar matching and ranking problems.

### 1.1   Related work

**Image geolocalization** uses recognition techniques from computer vision to estimate the location (at city, region, or global scale) of ordinary ground level photographs. Early work by Hays and Efros [12] studied the feasibility of this task by leveraging millions GPS-tagged images from the Internet. In [37], image localization is done efficiently by building a dataset of Google street-view images from which SIFT features are extracted, indexed and used for localization of a query image by voting. Lin et al. [19] propose the first ground-to-overhead geolocalization method. No attempt is made to learn a common feature space or match directly across views. Instead, ground-to-ground matching to a reference database of ground-overhead view pairs is used to predict the overhead features of a ground-level query. Bansal et al. [3] match street-level images to aerial images by proposing a feature which encodes facade structure self-similarity. Shan et al. [27] propose a fully automated system that registers ground-based multi-view Stereo models to aerial imagery using traditional multi-view and Structure from Motion technique.

**Deep learning** has been successfully applied to a wide range of computer vision tasks such as recognition of objects [17], places [39], faces [29]. Most recently, "PlaNet" [33] made use of a large amount of geo-tagged images, quantized the gps-coordinate into a number of regions and trained a CNN to classify an image's location into one of those regions. More relevant to this work is deep learning applications in cross-view images matching [20, 34]. The most similar published work to ours is Lin et al. [20] which uses a Siamese network to learn a common deep representation of street-view images and 45 degree aerial or bird's eye images. This representation is shown to be better than hand-crafted or off-the-shelf CNN features for matching buildings' facades from different angles. In [34], Workman et al. show that by learning different CNNs for different scales (i.e. using aerial images at certain scales), geolocalization can be done at the local or continental level. Interestingly, they also showed that by fixing the representation of ground-level image, which is 205 categories scores learned from the Places database [39], the CNN will learn the same category scores for aerial images. Most recently, Altwaijry et al. [2] use a deep attentive architecture to match aerial images across wide baselines.

## 2  Dataset of street view and overhead image pairs

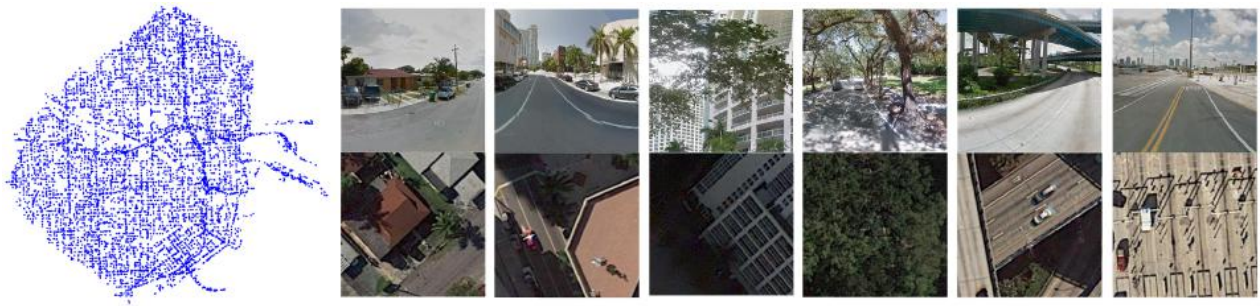

**Fig. 2.** On the left: visualization of the positions of all Miami's panorama images that we randomly collect for further processing. On the right: examples of produced street-view and overhead pairs.

We study the problem of matching street-view image to overhead images for the application of image geolocalization. To that end, we collect a large scale dataset of street-view and overhead images. More specifically, we randomly queried street-view panorama images from Google Map of the US. For each panorama, we randomly made several crops and for each crop we queried Google Map for the overhead image at the finest scale, resulting in an aligned pair of street-view and overhead images. Note that we want to localize the *scene* depicted in the image and not necessarily the camera. This is possible since Google panorama images come with geo-tags and depth estimates. We performed this data collection procedure on 11 different cities in the US and produced more than 1 million pairs of images. Some example matches in Miami are shown in Figure 2. We make this dataset available to the public.

Some similar attempts to collect a dataset for cross-view images matching task are [20,34], but neither are publicly available. We expect that the result and analysis here can be easily generalized across other datasets (or other applications like recognizing face or object instead of scene). While the technical aspects are similar, there will be qualitative differences: when training on [20], the network learns to match the facade which is visible from both views. On [34], the network learns to match similar categories of scenes or land cover types. And on our dataset, the network learns to recognize different fine-grained street scenes.

## 3   Cross-view matching and ranking with CNN

Before considering the ranking/retrieval task, we start with the matching/verification task formalized as following: during training phase, matched pairs of street-view and overhead images are provided as positive examples (negative examples can be easily generated by pairing up non-matched images) to learn a model. During testing, given a pair of images, the learned model is applied to classify if the pair is a match or not.

We use deep CNNs which have been shown to perform better than traditional hand-crafted features, especially for problems with significant training data available. We study 2 categories of CNNs (Figure 3): the classification network for recognizing matches and the representation learning networks for embedding cross-view images into the same feature space. Note that the first category is not practical for the large-scale retrieval application and is used as a loose upper bound for comparison.

The second category includes the popular Siamese-like network and the triplet network. We introduce another version of Siamese and triplet networks that use the *distance based logistic* layer, a novel loss function. For completeness we also include the Siamese-classification hybrid network (which will belong to the first category). In this section we will experiment with 6 networks in total.

### 3.1   Classification CNN for image matching

Since our task is basically classification, the first network we experiment with is AlexNet[17], originally demonstrated for object classification (Figure 3(a)). It has 5 convolutional layers, followed by 3 fully-connected layers and a soft-max

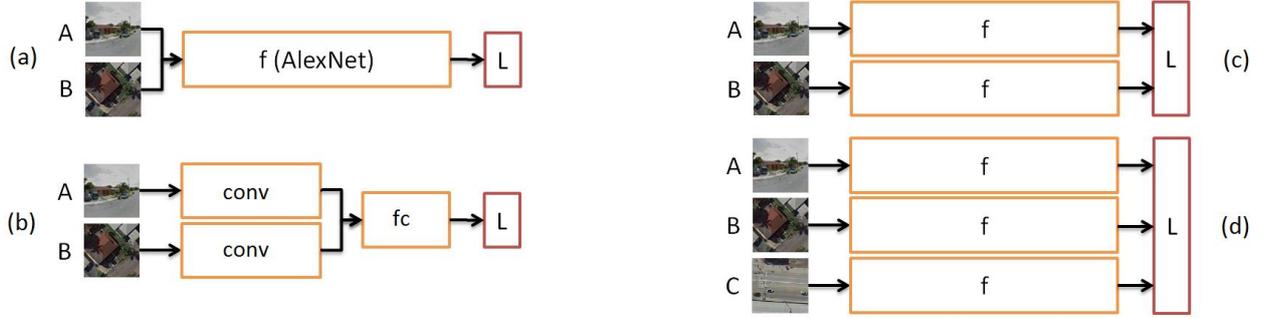

**Fig. 3.** Different CNN architectures: on the left is the first category: the classification network and the Siamese-classification hybrid network, on the right is the second category: the Siamese network and the triplet network

layer for classification. We make several modifications: (1) the input will be a 6-channel image, a concatenation of a street-view image and an overhead image, while the original AlexNet only takes 1-image input , (2) we double the number of filters in the first convolutional layer, (3) we remove the division of filters into 2 groups (this was done originally because of GPU memory limitation) and (4) the softmax layer produces 2 outputs instead of 1000 because our task is binary classification. Similar architectures have been used for comparing image patches [36].

Training the CNN is done by minimizing this loss function:

$$L(A, B, l) = LogLossSoftMax(f(I), l) \qquad (1)$$

where A and B are the 2 input images, $l \in \{0, 1\}$ is the label indicating if it's a match, $I = concatenation(A, B)$ and $f(.)$ is the AlexNet that outputs class scores.

### 3.2 Siamese-like CNN for learning image features

The Siamese-like network, shown in Figure 3(b), has been used for cross-view image matching [20, 34] and retrieval [30, 4]. It consists of 2 separate CNNs. Each subnetwork takes 1 image as input and output a feature vector. Formally, given 2 images A and B, we can apply the learned network to produce the representation f(A) and f(B) that can be used for matching. This is done by computing the distance between these 2 vectors and classifying it as a match if the distance is small enough. During training, the contrastive loss is used:

$$L(A, B, l) = l * D + (1 - l) * max(0, m - D) \qquad (2)$$

where D is the squared distance between f(A) and f(B), and m is the margin parameter that omits the penalization if the distance of non-matched pair is big enough. This loss function encourages the two features to be similar if the images are a match and separates them otherwise; this is visualized in Figure 4(left).

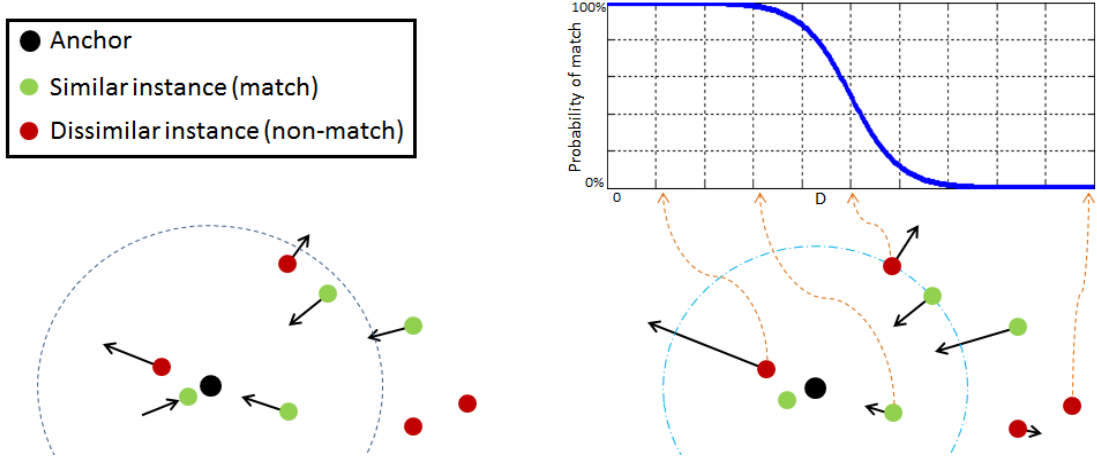

**Fig. 4.** Visualization of Siamese network training. We represent other instances (matches and non-matches) relative to a fixed instance (called the anchor). Left: with contrastive loss, matched instances keep being pulled closer, while non-matches are pushed away until they are out of the margin boundary, Right: log-loss with DBL: matched/nonmatched instances are pushed away from the "boundary" in the inward/outward direction.

In the original Siamese network [10], the subnetworks (f(A) and f(B)) have the same architecture and share weights. In our implementation, each subnetwork will be an AlexNet without weight sharing since the images are of different domains: one is street view and the other is overhead.

### 3.3 Siamese-classification hybrid network

The hybrid network is similar to the Siamese in that the input images are processed independently to produce output features and it is similar to the classification network that the features are concatenated to jointly infer the matching probability (Figure 3(c)). Similar architectures have been used for used for cross-view matching and feature learning [36, 1, 11, 2].

Formally let AlexNet ($f$) is consist of 2 parts: the set of convolutional layers ($f_{conv}$) and the set of fully-connected layers ($f_{fc}$), the loss function is:

$$L(A, B, l) = LogLossSoftMax(f_{fc}(I_{conv}), l) \quad (3)$$

Where $I_{conv} = concatenation(f_{conv}(A), f_{conv}(B))$. We expect this network to approach the accuracy of the classification network, while being slightly more efficient because intermediate features only need to be computed once per image.

### 3.4 Triplet network for learning image features

The fourth network that we call the triplet network or ranking network, shown in Figure 3(c), is popular for image feature learning and retrieval [35, 31, 24, 32, 26, 25], though its effectiveness has not been explored in cross-view image matching. More specifically it aims to learn a representation for ranking relevance between images. It consists of 3 separate CNNs instead of 2 in the Siamese network. Formally, the network takes 3 images A, B and C as inputs, where (A,B) is a match and (A,C) is not, and minimizes this hinge loss for triplet (which has been explored before its application in deep learning [6, 22]):

$$L(A, B, C) = max(0, m + D(A, B) - D(A, C)) \quad (4)$$

Where D is the squared distances between the features f(A), f(B), f(C), and m is the margin parameter to omit the penalization if the gap between 2 distances is big enough. This loss layer encourages the distance of the more relevant pair to be smaller than the less relevant pair (Figure 5(left)).

In the context of image matching, a pair of matched images (as the anchor and the match), plus a random image (as the non-match) is used as training example. With the learned representation, matching can be done by thresholding just like the Siamese network case.

### 3.5 Learning image representations with distance-based logistic loss

Despite being intuitive to understand, common loss functions based on euclidean distance might not be optimal for recognition. We instead advocate loss functions similar to the standard softmax, log-loss.

For the Siamese network, instead of the contrastive loss, we define the distance based logistic (DBL) layer for pairs of inputs as:

$$p(A, B) = \frac{1 + exp(-m)}{1 + exp(D - m)} \quad (5)$$

This outputs a value between 0 and 1, as the probability of the match given the squared distance. Then we can use the log-loss like the classification case for optimization:

$$L(A, B, l) = LogLoss(p(A, B), l) \quad (6)$$

The behavior of this loss is visualized in Figure 4(right). Notice the difference from the traditional contrastive loss.

For the triplet network, we define the DBL for triple as following:

$$p(A, B, C) = \frac{1}{1 + exp(D(A, B) - D(A, C))} \quad (7)$$

This represents the probability that it's a valid triple: B is more relevant to A than C is to A (note that $p(A, B, C) + p(A, C, B) = 1$). Similarly the log-loss function is used, so:

$$L(A, B, C) = log(1 + exp(D(A, B) - D(A, C))) \quad (8)$$

The behavior of this loss is visualized in Figure 5(right).

With this novel layer, we obtain Siamese and triplet DBL-Net that allow us to optimize for the recognition accuracy more directly. As with the original loss functions, the learned feature representation can be used for efficient matching and ranking at test time (when the DBL layer is not involved).

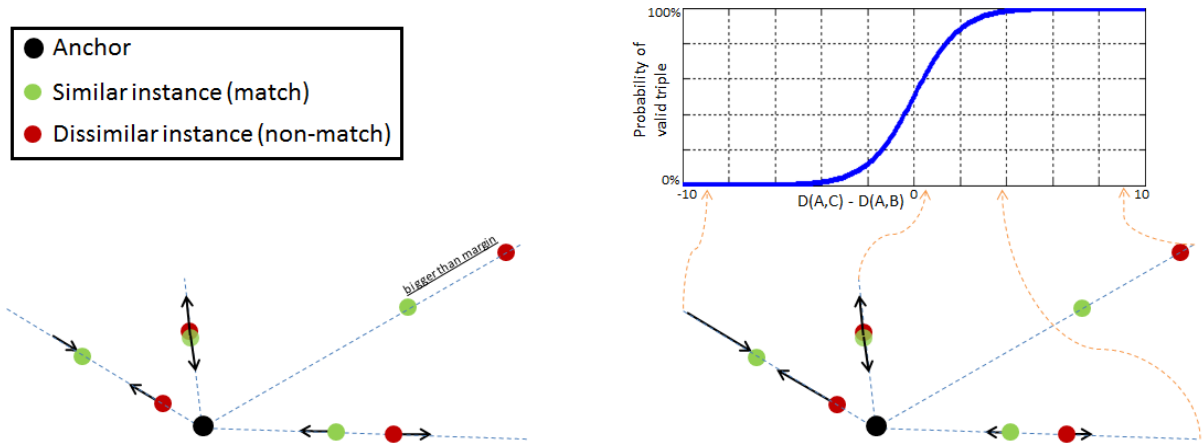

**Fig. 5.** Visualization of triplet network training. Each straight line originating from the anchor represents a triple. Left: with triplet/ranking loss, instances are pulled and pushed until the difference between the match distance and the non-match distance is bigger than the threshold, Right: log loss with DBL for triple. Similar to the ranking loss, but instead of relying on the threshold, the "force" depends on the current performance and confidence of the network.

**Implementation detail:** we use m=10; and D(.) is squared Euclidean distance. We do not do feature normalization (L2) in all of our experiment; hence the network can change the scale of the feature and the formulas here can be applied directly. However if there's normalization which basically predefined the scale of the output feature (and therefore the distance between them), it's best to scale the feature by a suitable constant (for example 3) before applying the DBL-log loss. Or equivalently change/validate the steepness and the midpoint of the logistic curve (instead of using the standard logistic function form).

## 4   Learning to perform rotation invariant matching

As we are considering the task of fine-grained street view to overhead view matching, not only spatial but also orientation alignment is important, i.e. rotating the overhead image according to the street-view's orientation instead of keeping the overhead image north oriented.

We aim to learn a rotation invariant (RI) representation of the overhead images. Similarly, Ke at al [16] studied the problem of shape recognition without explicit alignment. In [21], nearby filters are untied to potentially allow pooling on output of different filters. This helps to learn complex representation without big filters or increasing the number of filters; however that doesn't result in an explicit RI property like we desire. Deep symmetry network [9] is capable of encoding such a property, though its advantage is not significant when training data is sufficient for traditional CNN to learn that on its own. More relevant, [8] uses data augmentation and concatenation of features from different viewpoints.

However our training data comes with orientation aligned images (though not the test sets), which can potentially provide stronger supervision during training. In this section we explore techniques to take advantage of such information.

### 4.1   Partial rotation invariance by data augmentation

**Training with multiple rotation samples:** Rotation invariance (RI) can be encouraged simply by performing random rotation of overhead training images. Although invariance can help to a certain extent, there is a trade-off with discriminative ability. We propose to control the amount of rotation that the matching process will be invariant to, i.e. partial RI. Specifically this is done by adding a random amount of rotation within a certain range to the aligned overhead images. For example a 90° RI is achieved by rotating by an amount from −45° to 45°; 360° RI means fully RI.

**Testing with multiple rotation samples/crops:** since we don't know the correct orientation alignment at test time, if our representation is only partially rotation invariant, we have to test with multiple rotated version of the original image to find the best one. For example: with 360° RI representation, 1 sample is enough, with 180° RI representation, at least 2 rotation samples (that are 180° apart) are needed. Similar to multi-crop in classification tasks, we find that using more test time samples improves the result slightly (e.g. using 16 rotation samples at test time even if the network was trained to be 90° RI).

**Multi-orientation feature averaging:** as we use more rotation samples than needed, not only one but multiple of them should be good matches. For example testing with 16 rotation, we expect 16 of the them are good matches under 360° RI range, 4 under 90° RI range, etc. Therefore it makes sense to, instead of matching with a single best rotation (nearest neighbor), match with the best sequence of rotations. We propose to, depending on the degree of RI, average the features of multiple rotation samples during indexing time to obtain more stable features. This technique is especially useful in full RI case: all samples are averaged to produce a single feature, so the cost during query time is the same as using 1 sample.

### 4.2   Learning better representations with orientation regression

Next we propose to add an auxiliary loss function for orientation regression, where the amount of added rotation during training can be used as label for supervision. As shown in Figure 6, the features from the last hidden layer (fc7) are concatenated, then we add 2 fully connected layers (one acting as hidden layer and one as output layer) and use Euclidean distance as our loss function for regression.

It is known that additional or 'auxiliary' losses can be very useful. For example, ranking can be improved by adding a classification layer predicting category [4, 25] or attributes [14]. In [28], co-training of verification and classification is done to obtain a good representation for faces. Somewhat differently, our auxiliary loss is not directly related to the main task and its label is randomly generated by data augmentation. As the inference is done on 2 images jointly, its effect on each individual's representation can be difficult to interpret. The motivation, beyond being able to predict query orientation, is that this will make the network more orientation-aware and therefore produce a better feature representation for the localization task.

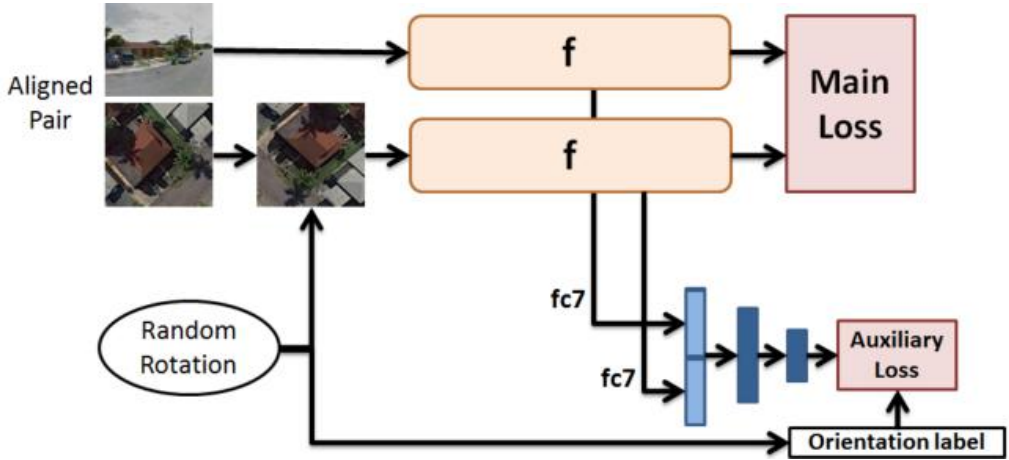

**Fig. 6.** Network architecture with data augmentation by random rotation and an additional branch that performs orientation regression

## 5   Experiments

Data preparation: we use our dataset of more than 1 million matched pairs of street-view and overhead-view images randomly collected from Google Maps of 11 different US cities (section 2). We use all the cross-view pairs in 8 cities as training data (a total of 900k examples) and the remaining 3 cities as 3 test sets (around 70k examples per set).

We learn with mini-batch stochastic gradient descent, the standard optimization technique for training deep networks. Our batch size is 128 (64 of which are positive examples while 64 are negative examples). Training starts with a large learning rate (experimentally chosen) and get smaller as the network converges. The number of training iterations is 150k. We use Caffe framework [15].

**Data augmentation:** we apply random rotation of overhead images during training and use multiple rotation samples during testing (described in Section 4). The effect will be studied in detail in section 5.2. We also apply a small amount of random cropping and random scaling.

**Image Ranking and Geolocalization**. While we have thus far considered location matching as a binary classification problem, our end goal is to use it for geolocalization. This application can be framed as a ranking or retrieval problem: given a query street view image and a repository of overhead images, one of which is the match, we want to rank the overhead images according to their relevance to the query so that the true match image is ranked as high as possible. The ranking task is typically approached as following: the representation learning networks are applied to the query image and the repository's images to obtain their feature vectors. Then these overhead images can be ranked by sorting the distance from their features to the query image's feature. The localization is considered successful if the true match overhead image is ranked within a certain top percentile.

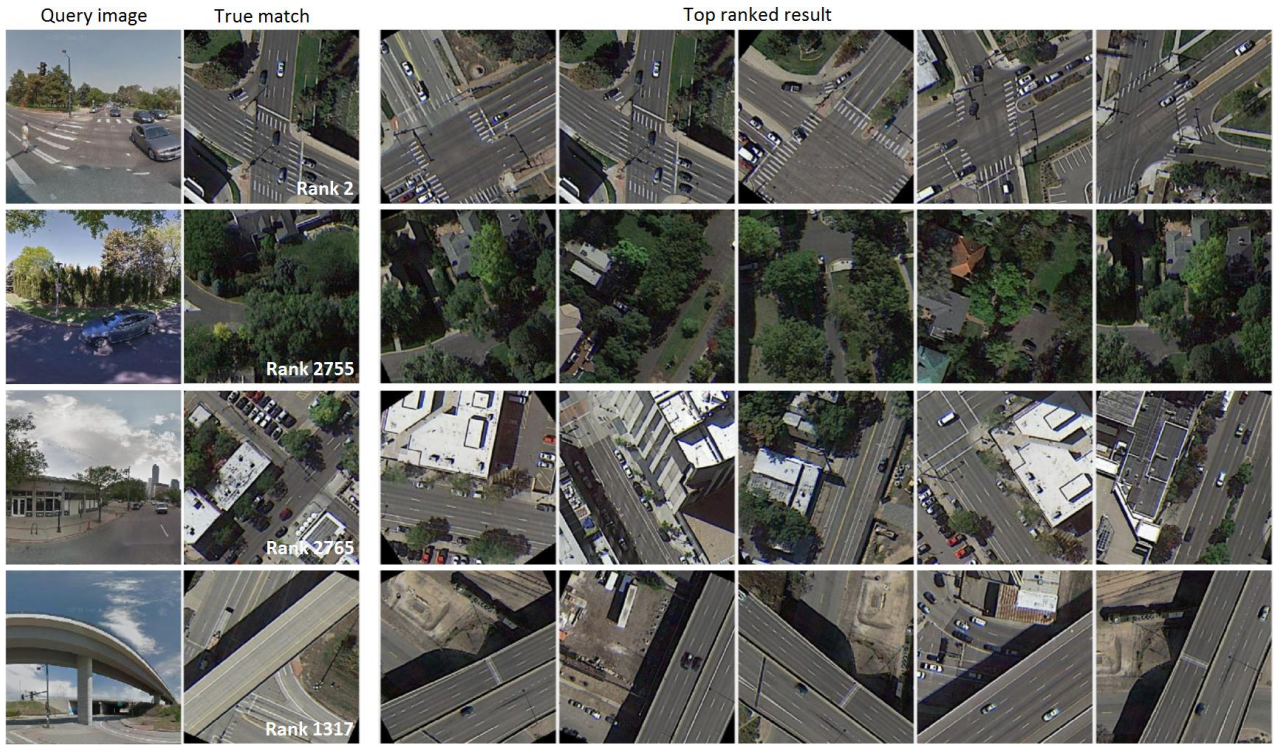

**Fig. 7.** Ranking result examples on the Denver test set (reference set of 70k reference images)

**Metrics:** we measure both the classification and ranking performance on each test set. The classification accuracy is computed by using the best threshold on the each test set (random chance performance is 50%). We found that this measurement is useful for evaluating classification networks which are hard to apply to ranking on large test sets because of the computational expense of all-pairs comparisons through deep networks. For the ranking task, we use mean recall at top K% as our measurement (the percentage of cases in which the correct overhead match of the query street view image is ranked within top K percentile, chance performance is K%). Some ranking examples are shown in Figure 7.

### 5.1  Comparison of CNN architectures

We train and compare 6 variants of CNN described in Section 3. All are initialized from scratch (no pretraining), trained to be 90° RI, and tested with 16 rotation samples. Quantitative comparisons are shown in the top of Table 1.

Not surprisingly, both classification networks achieved better accuracy than the representation learning Siamese and triplet networks. This is because they jointly extract and exchange information from both input images. Somewhat unexpectedly, in our experiments the hybrid network is the better of the two. Even-though the 'pure' classification network should be capable of producing the same mapping as the hybrid, it might have trouble learning to process both images from the 1st layer.

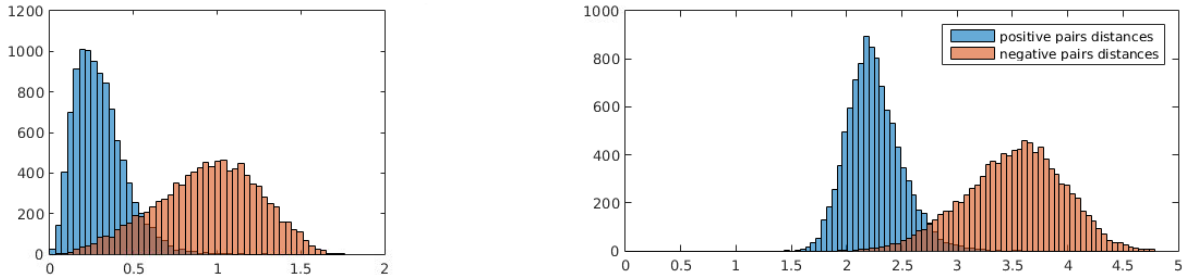

**Fig. 8.** Histograms of pairwise distances of features produced by the Siamese network-contrastive loss (left) and the triplet network (right). Note the crowding near zero distance for the Siamese network, which may explain poor performance for fine-grained retrieval tasks when it is important to compare small distances.

Between the Siamese and triplet network, the triplet network outperforms the Siamese by a surprisingly large margin on both tasks. While both networks try to separate matches from non-matches, the contrastive loss function works toward a secondary objective: drive the distance between matched pair as close to 0 as possible (Figure 8). Note that this might be a good property for the learned representation to have; but for the task of matching and ranking we found that this might compromise the main objective. One way to alleviate this problem is to add another margin to the contrastive loss function to cut the loss when the distance is small enough [18].

**Table 1.** Performance of different networks on different test sets

| Task | Classification (accuracy) | | | Ranking (recall @top 1%) | | |
|---|---|---|---|---|---|---|
| Test set | Denver | Detroit | Seattle | Denver | Detroit | Seattle |
| Section 5.1 experiment (90°RI+16rots) | | | | | | |
| Classification network | 90.0 | 87.8 | 87.7 | N/A | N/A | N/A |
| Classification hybrid | 91.5 | 88.7 | 89.4 | N/A | N/A | N/A |
| Siamese network | 85.6 | 83.2 | 82.9 | 21.6 | 21.9 | 17.7 |
| Triplet network | 88.8 | 86.8 | 86.4 | 43.2 | 39.5 | 35.3 |
| Siamese DBL-Net | 90.0 | 88.0 | **88.0** | 48.4 | 45.0 | **41.8** |
| Triplet DBL-Net | **90.2** | **88.4** | 87.6 | **49.3** | **47.1** | 40.0 |
| Section 5.2 (360°RI+OR) | | | | | | |
| DBL-Net + 16rots | **91.5** | **90.1** | 88.7 | **54.8** | **52.7** | **45.5** |
| DBL-Net + avg16 | **91.5** | 90.0 | **88.8** | 54.0 | 52.2 | 45.3 |
| Section 5.3 | | | | | | |
| Triplet eDBL-Net | **91.7** | **89.9** | **89.3** | **59.9** | **57.8** | **51.4** |

Analysis of Siamese and triplet network's performance has helped us develop the DBL layer. As the result, both DBL-Nets significantly outperform the original networks. While the Siamese with DBL and triplet network with DBL have comparable performances, it seems that the triplet DBL-Net is slightly better at ranking. Note that for most of the experiments we have been conducting, the performance of these two tasks strongly correlate. We use the triplet network with DBL layer for all following experiments.

### 5.2  Rotation invariance

We experiment with partial rotation invariance (RI) and orientation regression (OR) (described in Section 4) for matching and ranking using the triplet DBL-Net. The result is shown in Table 2.

**Table 2.** Comparisons of different amount of partial rotation invariance (RI), with and without orientation regression (OR), and different numbers of rotation samples during test time. In this experiment, the triplet network with DBL layer is tested on the Denver test set. 1GT*: in this setting, we test with 1 overhead image aligned using the ground-truth orientation (so the network doesn't have to be RI).

| Task | Classification (accuracy) | | | | Ranking (recall @top 1%) | | | |
|---|---|---|---|---|---|---|---|---|
| Number of test rotations | 1 | 4 | 16 | 1GT* | 1 | 4 | 16 | 1GT* |
| 0° RI (no RI) | 63.6 | 68.5 | 87.2 | 95.0 | 11.0 | 18.8 | 37.3 | 76.2 |
| 45° RI | 70.9 | 86.2 | 89.9 | N/A | 19.3 | 36.8 | 48.1 | N/A |
| 90° RI | 75.8 | 89.5 | **90.2** | N/A | 24.7 | 44.7 | **49.3** | N/A |
| 180° RI | 82.7 | 89.2 | 89.6 | N/A | 31.2 | 43.0 | 45.6 | N/A |
| 360° RI (full RI) | 87.7 | 88.5 | 88.9 | N/A | 36.8 | 40.0 | 41.9 | N/A |
| 90° RI + OR | 74.3 | 88.6 | 89.4 | N/A | 23.1 | 43.4 | 47.4 | N/A |
| 360° RI + OR | 90.9 | 91.3 | **91.5** | N/A | 50.9 | 53.2 | **54.8** | N/A |
| 360° RI + OR + avg16 | **91.5** | N/A | N/A | N/A | 54.0 | N/A | N/A | N/A |

As an upper bound, we train a network where overhead images are aligned to the ground truth camera direction of the street view image (1GT). This is not a realistic usage scenario for image geolocalization since camera azimuth would typically be unknown. As expected, the network without RI performs very well when true alignment is provided during testing (1GT), but performs poorly otherwise. This baseline shows how challenging the problem has become because of orientation ambiguity. As the degree of RI during training is increased, the performance improves.

Observe that fewer numbers of test time rotated crops/samples doesn't work well if the amount of RI is limited. The full RI setting is the best when testing with a single sample. As the number of rotations increase, the performance improves, especially for the partially RI networks. Using 16 rotations, the 90° RI network has the highest performance. It might be the best setting for compromising between invariance and discriminate power (this might not be the case when using hundreds of samples, but we found that it's not computationally practical and the improvement is not significant).

Orientation regression's impact on the 360° RI network is surprisingly significant; its performance improves by 30% (relatively). However OR doesn't affect 90° RI network positively, suggesting that the 2 techniques might not complement each other. It's interesting that the OR is useful even though its effect during learning is not as intuitive to understand as partial RI. As a by-product, the network can align matches. The orientation prediction has an average error of 17° for the ground truth matching overhead image and is discussed more in the supplemental document.

Finally we show the effect of applying multi-orientation feature averaging on 360° RI + OR network. By averaging the feature of 16 samples, we obtain comparable performance to exhaustively testing with 16 samples (result on all 3 test sets is shown in the 2nd part of Table 1). Though not shown here, applying this strategy to partial RI networks also slightly improve their performances.

### 5.3   Triplet sampling by exhausting mini-batch

To speed up the training of triplet networks with the triplet hinge loss, clever triplet sampling and hard negative mining is usually applied [32, 26, 31]. This is because the triplet not violating the margin does not contribute to the learning. However it can skew the input distribution if not handled carefully (for instance, only mine hardest examples); different schemes were used in [32, 26, 31].

On the other hand, our DBL-log loss is practically a smoothed version of the hinge loss. We propose to use every possible triplet in the mini-batch. We experiment with using a mini-batch of 128 pairs of (matched) images. Since each image in our data has a single unique match only, we can generate a total of 256 * 127 triplets (256 different anchors, 1 match and 127 non-matches per anchor). This is done within our exhausting DBL log loss layer implementation (eDBL); hence the cost of processing the mini-batch is not much more expensive. In a similar spirit, recent work[23] proposes a loss function that considers the relationship between every examples in each training batch.

We train a triplet eDBL-Net+360°RI+OR+avg16. Its effect is very positive: the convergence is much faster, after around 30k iterations the network achieved similar performance as in previous experiments where each network was trained with 150k iterations using the same batch size. After 80k iterations, we achieve even better ranking performance, shown at the bottom of table 1.

## 6   Conclusion

We introduce a new large scale cross-view data of street scenes from ground level and overhead. On this dataset, we have experimented with different CNN architectures extensively; the reported results and analysis can be generalized to other ranking and embedding problems. The result indicates that the Siamese network with contrastive loss is the least competitive even though it has been popular for cross-view matching. Our proposed DBL layer has significantly improved representation learning networks. Last but not least, we show how to further improve ranking performance by incorporating supervised alignment information to learn a rotational invariant representation.

**Acknowledgments.** Supported by the Intelligence Advanced Research Projects Activity (IARPA) via Air Force Research Laboratory, contract FA8650-12-C-7212. The U.S. Government is authorized to reproduce and distribute reprints for Governmental purposes notwithstanding any copyright annotation thereon. Disclaimer: The views and conclusions contained herein are those of the authors and should not be interpreted as necessarily representing the official policies or endorsements, either expressed or implied, of IARPA, AFRL, or the U.S. Government.

# Supplemental Material: Localizing and Orienting Street Views Using Overhead Imagery

## 1  GTCrossView dataset

Figure 1 shows the cities from which we collected data (initially we wanted to use both big and small city/town, but the image quality seems to be quite inconsistent).

Majority of the images in the dataset are of rural-like scene because the urban area is relatively small even in big cities.

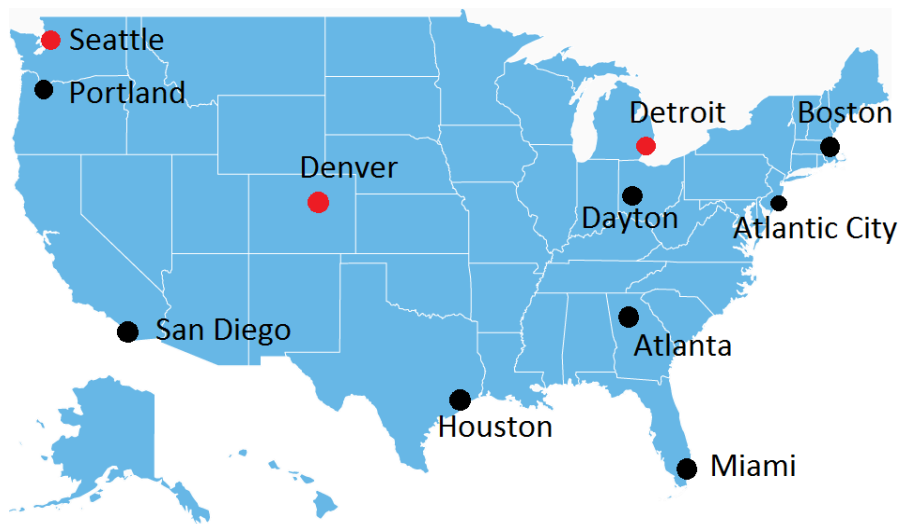

**Fig. 1.** Location of cities we chose to build our dataset. Black: we use for training, and Red: we use for testing in our experiments.

## 2  Network visualization

### 2.1  conv1

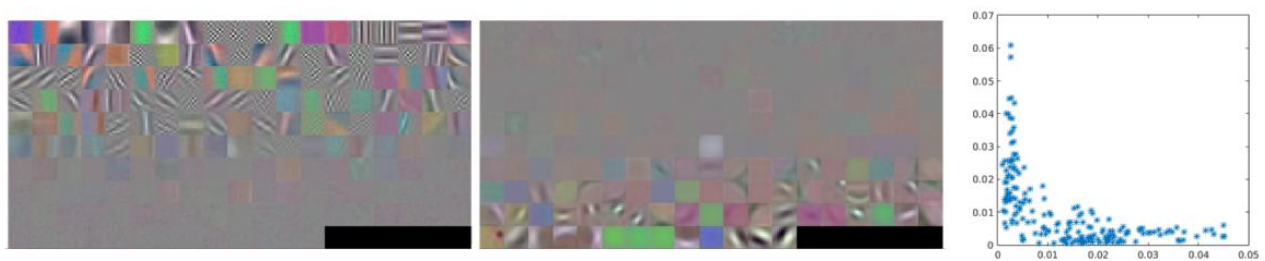

**Fig. 2.** conv1 filters learned by the classification network

First we visualize the first convolutional layer (named conv1 in AlexNet) learned by these networks. Figure 2 shows conv1 of the learned classification AlexNet. Since the input of our (modified) network is a 6-channel image, each convolutional filter also has 6 channels. We split it into 2 parts: channel 1-3 weights (which apply to street-view image) and channel 4-6 weights (which apply to overhead image).

A quick observation is that most the weights are (noisy) zero (gray-color) in either channel 1-3 or 4-6. This indicates that even though this network can combine and exchange information between 2 images, most filters in the first layer only focus on extracting feature from 1 image only. To be sure, we compute the standard deviation of channel 1-3 weights and channel 4-6 weights of each filter and plot all of them in figure 2-right. Most filters have 1 std higher than the others and none has both high std value. Another observation is that there's more filters focusing on street-view image than overhead image. In fact, the number of filters having higher channels 1-3 weights std is 111 (out of 192). One explanation is that the scenes in street-view images have greater variation than that of overhead images, hence needing more filters' focus to learn.

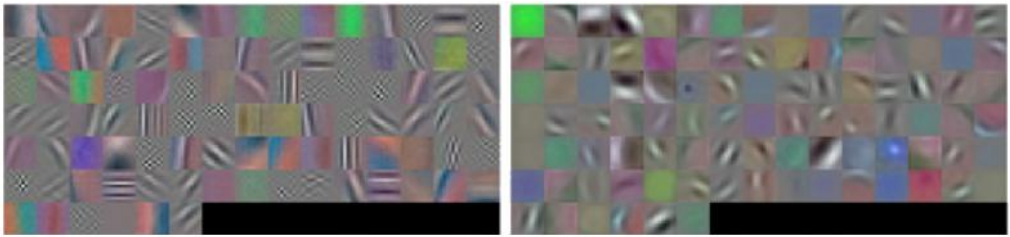

**Fig. 3.** conv1 filters learned by the representation learning network

In figure 3 we show conv1 filters learned by a representation learning networks are similar. Notice the difference between filters of the street-view image and overhead image. These filters are similar to their counterpart in the classification network.

### 2.2   Output feature activation

It's difficult to visualize the features learned in the other layers. In object recognition, they usually detect similar objects or objects' parts [38]. In cross-view image matching, they detect buildings with similar structural patterns [20]. Our features from the classification network learn to detect similar scenes (or pairs of scenes, in case of classification network). Figure 4 shows some images with extreme big value of an output feature in first 5 columns, and images with extreme small value of that same feature in the last 5 columns.

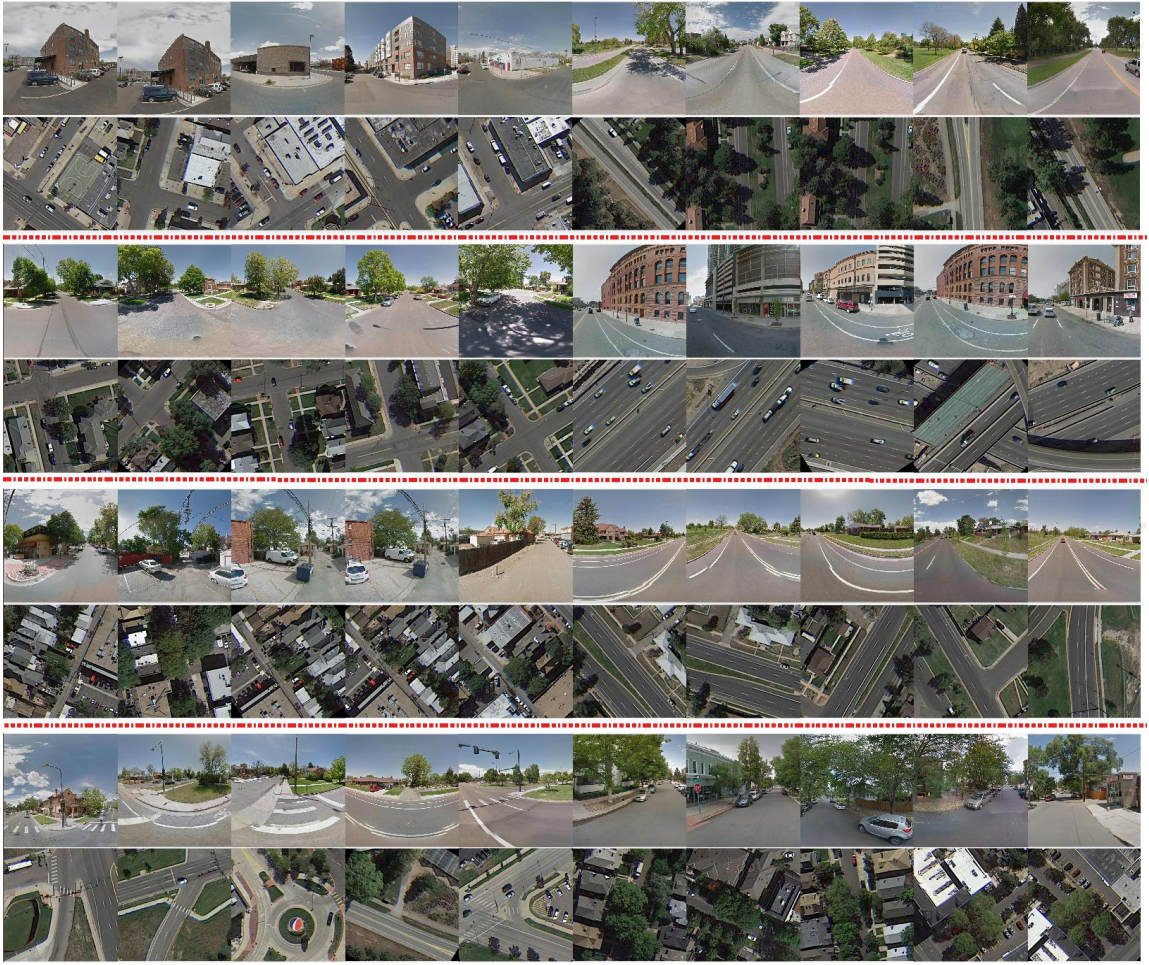

**Fig. 4.** Examples of images with extreme activation value

## 3  Ranking performance

Figure 5 shows the ranking performance of some networks that we have experimented with (described in the experiment section). The Siamese network baseline doesn't perform well relatively suggesting it is not suitable for ranking application. Each of our proposals (DBL + IR + OR + mini-batch exhausting) helps to improve the triplet network significantly.

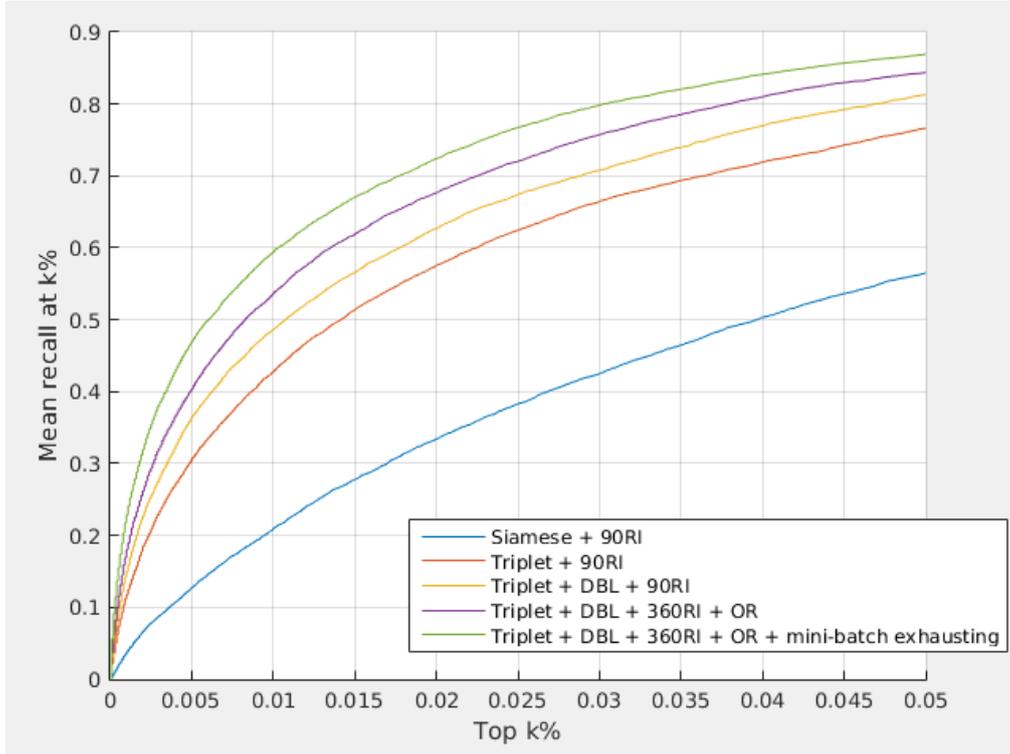

**Fig. 5.** Ranking performance on Denver test set

In figure 6 we show some image geolocalization examples. Assuming the position of the overhead images' scenes is known, we can infer the likely position of the scene in streetview image.

### 3.1  Comparison on other datasets (to be updated)

Most state-of-the-arts for verification or ranking using Siamese or triplet network has been demonstrated to be superior to using shallow or pre-trained features. Our DBL can help to further improve the performance.

We experimented on a smaller scale cross-view dataset from [20]. The dataset has around 80k pairs of matched street view images and aerial images in 7 cities; the task is to train on 31k pairs and test the ranking performance on the rest.

We train triplet network and triplet DBL-Net, both initialized from scratch. With mini-batch exhausting, the network fits really fast and begins to overfit after only 5k iterations (batch size: 32 pairs). To deal with that we apply heavy random cropping and random rotation within 10 degree. We run the training for 30k iterations; the result is shown in table 1. Triplet eDBL-Net seems to outperform [20] and traditional triplet network on most test sets (though it might not be directly comparable because our training data is slightly smaller than what has been originally used in [20]).

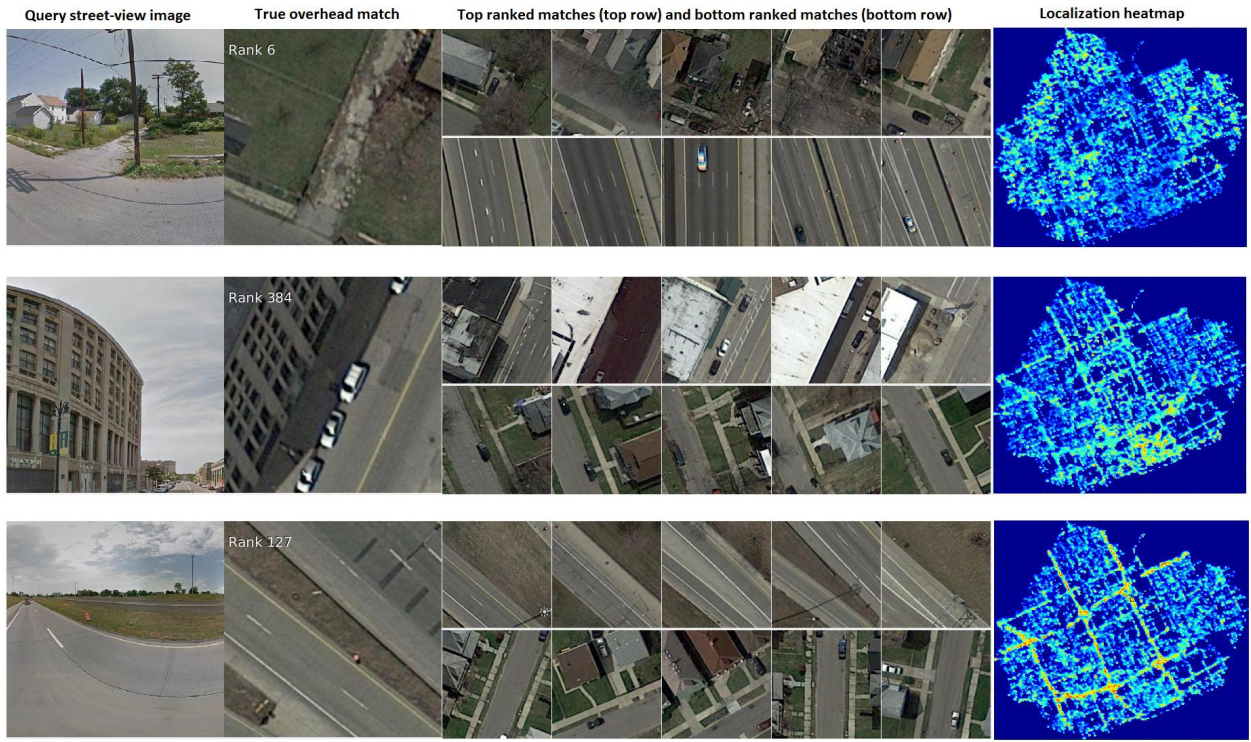

**Fig. 6.** 3 geolocalization examples on the Detroit city test set (85,345 overhead-view images)

Table 1. Ranking performance (Recall at 1%) on [20]'s dataset

| Test set | SF | Charleston | Chicago | SD | Tokyo | Rome | Lyon |
|---|---|---|---|---|---|---|---|
| Siamese [20] | 22.4 | 22.6 | 8.6 | 23.2 | 7.3 | 13.0 | **11.7** |
| Triplet (e-)Net | 26.0 | 33.1 | 12.3 | 24.5 | 7.3 | 13.6 | 8.5 |
| Triplet eDBL-Net | **33.8** | **40.8** | **18.2** | **32.0** | **10.2** | **17.2** | 11.1 |

## 4  Orientation regression performance

Our network with auxiliary OR loss is capable of predicting the orientation difference between street-view image and overhead-view image; though it's only a by-product and not used for our image geo-localization application. Here we report the network's performance on orientation prediction.

We compute the difference (in degree) between the true orientation and the predicted orientation; the average error (absolute difference) is around 17°. We plot the histogram of these differences on the Denver test set in figure 7. Notice most fall close to 0°, but there's a very small peak around −180/180°. This represents cases in which the scene looks symmetrical from aerial view point. We show some examples prediction in figure 8.

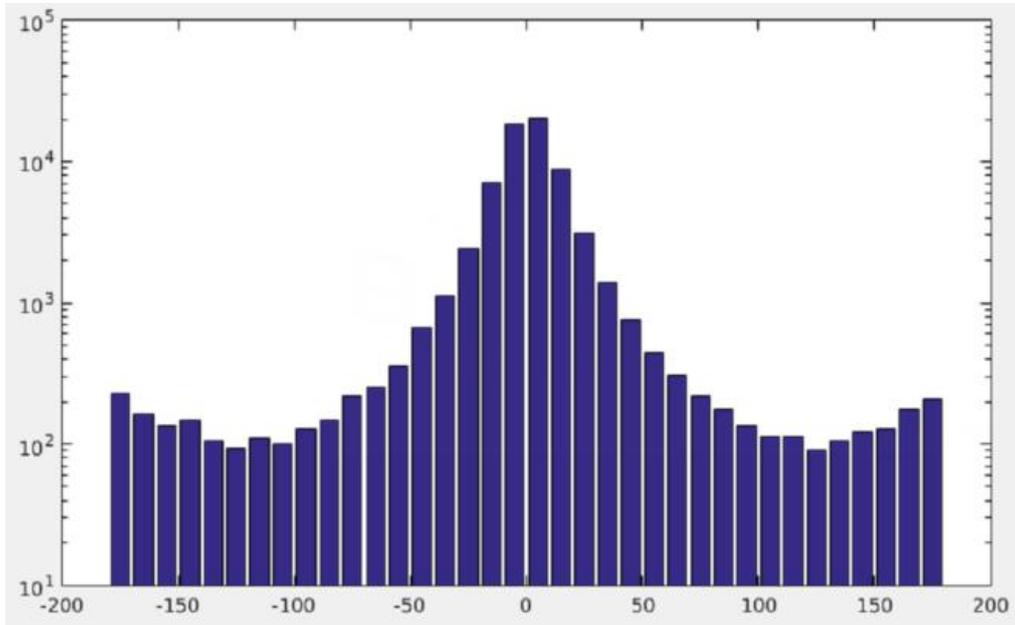

**Fig. 7.** Histograms of difference between predicted orientation and true orientation.

## 5   Residual network

One can benefit from using deeper network. Here we train a ResNet-101 model [13] and compare it with AlexNet version, the result is shown in table 2

**Table 2.** Compare AlexNet vs ResNet-101

| Task | Classification (accuracy) | | | Ranking (recall @top 1%) | | |
| --- | --- | --- | --- | --- | --- | --- |
| Test set | Denver | Detroit | Seattle | Denver | Detroit | Seattle |
| eDBL-AlexNet | 91.7 | 89.9 | 89.3 | 59.9 | 57.8 | 51.4 |
| eDBL-ResNet-101 | **92.4** | **91.5** | **91.5** | **60.7** | **64.0** | **58.4** |

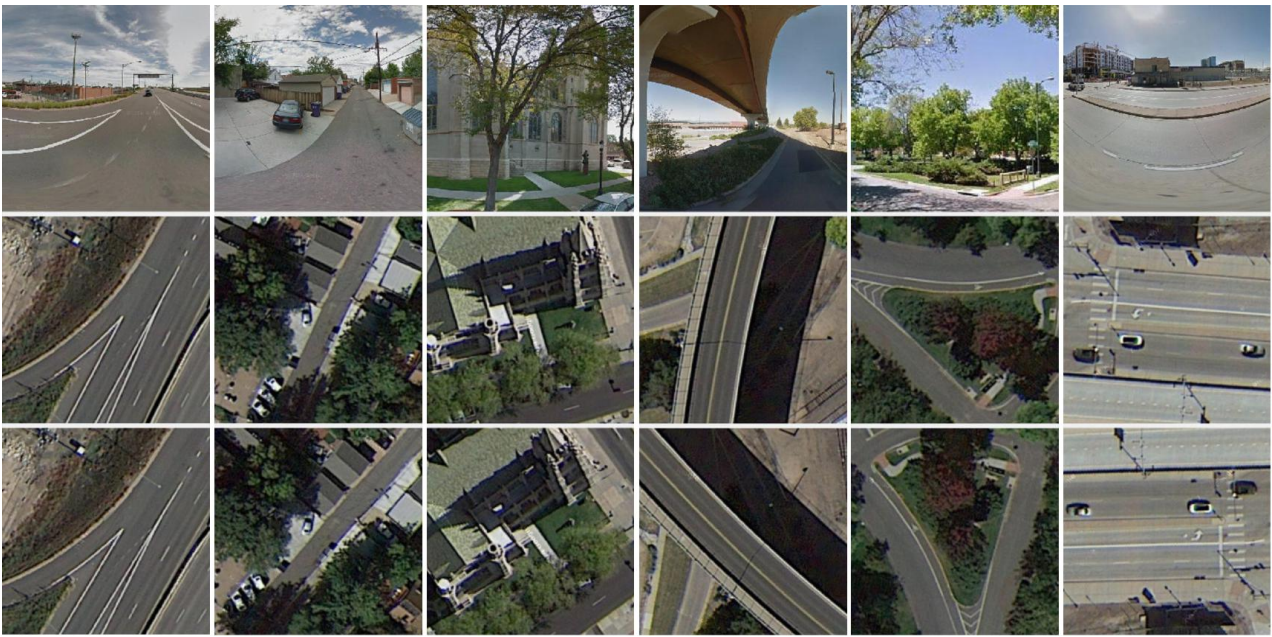

**Fig. 8.** Orientation prediction examples: first row is the street-view images, second row is the ground-truth aligned overhead images and third row is the alignments using predicted orientation.